\documentclass[letterpaper]{article} % DO NOT CHANGE THIS
\usepackage{aaai23}
\usepackage{times}  % DO NOT CHANGE THIS
\usepackage{helvet}  % DO NOT CHANGE THIS
\usepackage{courier}  % DO NOT CHANGE THIS
\usepackage[hyphens]{url}  % DO NOT CHANGE THIS
\usepackage{graphicx} % DO NOT CHANGE THIS
\usepackage{natbib}  % DO NOT CHANGE THIS AND DO NOT ADD ANY OPTIONS TO IT
\usepackage{caption} % DO NOT CHANGE THIS AND DO NOT ADD ANY OPTIONS TO IT
\usepackage{algorithm}
\usepackage{algorithmic}
\usepackage{graphicx}
\usepackage{amsmath}
\usepackage{amssymb}
\usepackage{color}
\usepackage{multirow}
\usepackage{amsthm}
\usepackage{amsfonts}
\usepackage{newfloat}
\usepackage{listings}
\DeclareCaptionStyle{ruled}{labelfont=normalfont,labelsep=colon,strut=off} % DO NOT CHANGE THIS
\floatstyle{ruled}
\newfloat{listing}{tb}{lst}{}
\floatname{listing}{Listing}
\title{Exploring Depth Information for Face Manipulation Detection}
\author {
    Haoyue Wang,
    Meiling Li,
    Sheng Li\thanks{*Corresponding author},
    Zhenxing Qian,
    Xinpeng Zhang
}
\affiliations {
    School of Computer Science, Fudan University\\
    \{20210240261, mlli20, lisheng, zxqian, zhangxinpeng\}@fudan.edu.cn
}

\usepackage{bibentry}
\begin{document}

\maketitle
\begin{abstract}
Face manipulation detection has been receiving a lot of attention for the reliability and security of the face images. Recent studies focus on using auxiliary information or prior knowledge to capture robust manipulation traces, which are shown to be promising. As one of the important face features, the face depth map, which has shown to be effective in other areas such as the face recognition or face detection, is unfortunately paid little attention to in literature for detecting the manipulated face images. In this paper, we explore the possibility of incorporating the face depth map as auxiliary information to tackle the problem of face manipulation detection in real world applications. To this end, we first propose a Face Depth Map Transformer (FDMT) to estimate the face depth map patch by patch from a RGB face image, which is able to capture the local depth anomaly created due to manipulation. The estimated face depth map is then considered as auxiliary information to be integrated with the backbone features using a Multi-head Depth Attention (MDA) mechanism that is newly designed. Various experiments demonstrate the advantage of our proposed method for face manipulation detection.   

\end{abstract}

\section{Introduction}

The development of deep learning techniques have made face manipulation an easy task. People can manipulate the face images using a variety of deepfake schemes \cite{thies2019deferred, thies2016face2face, perov2020deepfacelab, perez2003poisson, li2019faceshifter, huang2020fakepolisher}. The manipulated face images are usually difficult to be distinguished by human eyes, which seriously challenges the reliability and security of face images. It is of paramount importance to develop advanced and accurate face manipulation detection schemes.

Researchers have devoted a lot of efforts to the task of face manipulation detection. Various deep neural networks (DNN) are proposed to spot the difference between real and manipulated face images, such as ResNet \cite{he2016resnet}, Xception \cite{rossler2019faceforensics++}, MesoNet \cite{afchar2018mesonet}, and EfficientNet \cite{tan2019efficientnet}. Recently, researchers start to explore different auxiliary information or prior knowledge to facilitate the face manipulation detection, including the blending boundary \cite{li2020face}, identity \cite{cozzolino2021id}, pre-generated face attention mask \cite{zi2020wilddeepfake, Schwarcz2021finding}, face information in the frequency domain \cite{gu2021exploiting, chen2021local}, and the face texture features \cite{zhu2021face,zhao2021multi}. Such a strategy is shown to be promising for performance boosting.

Despite the progress that have been made, the performance of the existing schemes usually rely on the rich features that can be extracted from high quality fake face images (i.e., the high quality compressed or raw face images), where the details of the manipulation traces are available. In real-world applications, however, the classifiers learnt from the detailed manipulation traces in one dataset may not be robust against those from another dataset, which results in severe performance reduction in cross-database scenarios. On the other hand, the face images posted on the social networks may be heavily compressed for storage saving. The compression would eliminate the details of the image content as well as the manipulation traces, which also challenges the effectiveness of the existing schemes.  

A few attempts have been made in literature to propose schemes that are specific for the detection of low quality compressed manipulated face images. In \cite{woo2022add}, the authors take advantage of knowledge distillation to address such an issue, which relies on a teacher model trained on a large amount of high quality face images.

\begin{figure}%[htbp]
    \centering
    \includegraphics[width=0.35\textwidth]{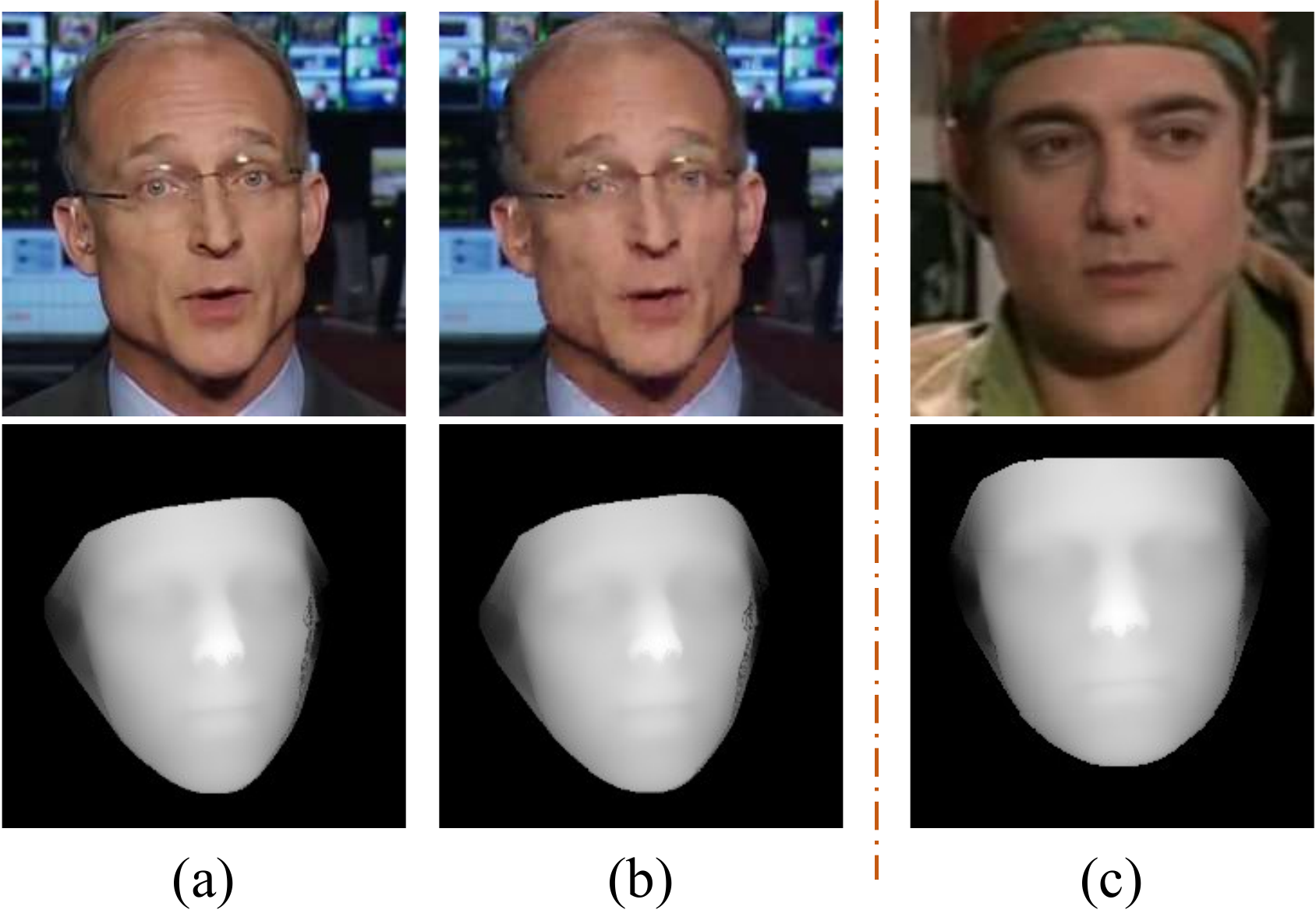}
    \caption{Face depth maps estimated from face images with different compression qualities and different sources. The face images in (a) and (b) are selected from FaceForensics++ \cite{rossler2019faceforensics++} with compression qualities of c23 and c40, the face image in (c) is selected from Celeb-DF \cite{li2020celebdf}. }
    \label{fig:depthmap_intuition}
\end{figure}

In this paper, we explore the possibility to estimate and incorporate the depth map of the face images for face manipulation detection. The rationales behind are these: 
\begin{itemize}
    \item The face depth map is not sensitive to image compression. As shown in Fig. \ref{fig:depthmap_intuition} (a) and (b), a lot of details in the face image are lost due to severe compression, however, the depth map features are not seriously affected, where we adopt the PRNet \cite{feng2018joint} for face depth map estimation.
    \item The face depth map tends to be stable among the face images that are collected from different sources (see Fig. \ref{fig:depthmap_intuition} (a) and (c)), while the manipulation would distort the face depth maps. Take the popular generative DNN based face manipulation as an example, the fake face region will either have no depth (if it is computer-generated) or have abnormal depth features around the boundary (if it is swapped from another real face image).    
\end{itemize}

There have been studies to estimate the depth map from a two dimensional RGB face image \cite{feng2018joint, jin2021face, wu2021dual, kang2021facial}, however, all these schemes assume the face image is real and captured from a human face which could be considered as a physical object with relatively smooth surface. The estimated face depth maps are globally smooth, which are not sensitive to face manipulation operations. To deal with this issue, we propose a Face Depth Map Transformer (FDMT) for face depth estimation, which is capable of capturing the local patch-wise depth variations due to the face manipulation. We further propose a Multi-head Depth Attention (MDA) mechanism to effectively incorporate the face depth map into the backbones for face manipulation detection. The proposed method is shown to be significantly better than the existing schemes in the cross-database scenario, which also achieves good performance in the intra-database scenario for detecting low quality compressed fake face images. We test the generalization ability of the proposed method on three popular face manipulation detection backbones including Xception \cite{chollet2017xception}, ResNet50 \cite{he2016resnet} and EfficientNet \cite{tan2019efficientnet}, all demonstrate the effectiveness of the proposed method for face manipulation detection. The contributions of this paper are summarized below.

\begin{itemize}
    \item We explore the possibility of using the face depth map, which is seldom considered in the area of face manipulation detection, for performance boosting.
    \item We propose a Face Depth Map Transformer (FDMT) for generating local patch-wise depth features that are sensitive to face manipulation.    
    \item We propose a Multi-head Depth Attention (MDA) mechanism to effectively integrate our face depth maps into different backbones for face manipulation detection. 
 \end{itemize}

\section{Related Works}

In this section, we briefly review recent works on face manipulation detection and face depth map estimation.

\subsection{Face Manipulation Detection}

With the continuous development of deep learning, various DNN backbones have been proposed for face manipulation detection. Wang \textit{et al.} \cite{wang2020cnn} apply ResNet \cite{he2016resnet} to classify real or fake face images. Rossler \textit{et al.} \cite{rossler2019faceforensics++} use Xception \cite{chollet2017xception} as a baseline DNN model, which can achieve satisfactory performance on intra-database evaluations. Afchar \textit{et al.} \cite{afchar2018mesonet} design a compact network MesoNet for video based face manipulation detection. Zhao \textit{et al.} \cite{zhao2021multi} propose to take advantage of the EfficientNet \cite{tan2019efficientnet} for face manipulation detection, which can achieve comparable performance to Xception. There are also patch-based face manipulation detection approaches proposed to extract the subtle manipulated traces located in the image patches. Chai \textit{et al.} \cite{chai2020makes} take advantage of a patch-based classifier with limited receptive fields in the image. The works in \cite{chen2021local, zhao2021learning} further consider the patch similarity to facilitate the classification, where each patch is equally treated and processed during the patch feature learning. Zhang \textit{et al.} \cite{zhang2022patch} propose a Patch Diffusion (PD) module to fully exploit the patch discrepancy for effective feature learning.

On the other hand, a lot of recent studies focus on exploring effective auxiliary information or prior knowledge for the task of face manipulation detection, which are shown to be promising. Li \textit{et al.} \cite{li2020face} take the facial blending boundary as an indicator for the existence of manipulation. Dang \textit{et al.} \cite{dang2020detection} propose to incorporate the position of the face manipulation area to make the network focus on the manipulation traces. Zi \textit{et al.} \cite{zi2020wilddeepfake} extract and fuse the face mask and organ mask into an attention mask to make the network pay attention to the fake area. Schwarcz \textit{et al.} \cite{Schwarcz2021finding} generate masks of the important parts of the face image to perform multi-part detection. The works in \cite{zhu2021face,zhao2021multi} introduce different approaches to extract the face texture features to guide the network for better detection of the manipulation cues. Masi \textit{et al.} \cite{masi2020two} adopt a dual-branch network structure, one of which is a fixed filter bank to extract the face feature in the frequency domain for auxiliary information. Similarly, the works in \cite{frank2020leveraging,qian2020thinking, gu2021exploiting, chen2021local} propose different approaches to treat the frequency domain face information as auxiliary information to boost the performance. Cozzolino \textit{et al.} \cite{cozzolino2021id} propose to use the identity information for detecting manipulated face videos, which requires a real face video for reference to conduct the detection.

Relatively few works are designed specifically for low quality compressed fake face image detection. Woo \textit{et al.}  \cite{woo2022add} propose to use knowledge distillation to transfer the knowledge learnt in a teacher model into a student model, where the teacher model is trained on high-quality face images.
%ADD 原文中使用high-quality raw images

\begin{figure*}[htbp]
    \centering
    \includegraphics[width=0.90\textwidth]{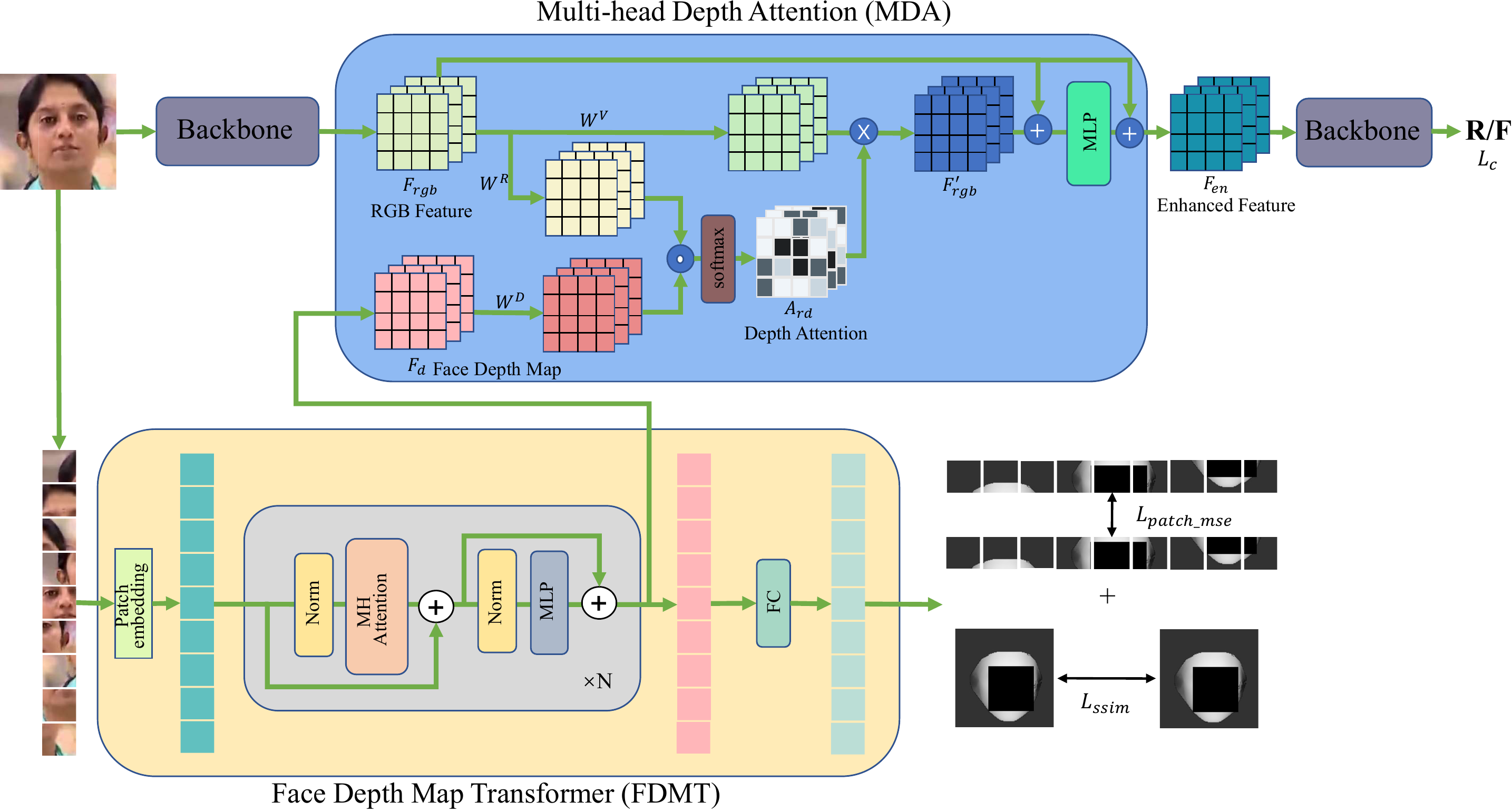}
    \caption{An overview of the proposed method for face manipulation detection.}
    \label{fig:framework}
\end{figure*}

\subsection{Face Depth Map Estimation}

In general, face depth map estimation tries to construct the depth information from one or more two dimensional RGB face images. Feng \textit{et al.} \cite{feng2018joint} propose a Position map Regression Network (PRNet) to exquisitely estimate the face depth. Since the scene depth is relatively easier to obtain compared with the face depth map, Jin \textit{et al.} \cite{jin2021face} apply scene depth knowledge for face depth map estimation. Wu \textit{et al.} \cite{wu2021dual} propose a depth uncertainty module to learn a face depth distribution instead of a fixed depth value. Kang \textit{et al.} \cite{kang2021facial} propose a StereoDPNet to perform depth map estimation from dual pixel face images.

The face depth map is shown to be a good feature for face related machine learning tasks. Chiu \textit{et al.}  \cite{chiu2021high} develop a segmentation-aware depth estimation network, DepthNet, to estimate depth maps from RGB face images for accurate face detection. Wang \textit{et al.} \cite{wang2020deep} argue that the depth map can reflect discriminative clues between live and spoofed faces, where the PRNet \cite{feng2018joint} is adopted to estimate the face depth map for face anti-spoofing. Zheng \textit{et al.} \cite{zheng2021attention} also take the face depth map as auxiliary information for face anti-spoofing, where a symmetry loss is proposed for reliable face depth estimation. 

Motivated by the effectiveness of the face depth map in the aforementioned applications, we believe it is worthy of investigation to see how we could take advantage of such information for face manipulation detection. We think the face depth map is a robust feature against compression and different capturing devices, which could be helpful when we are encountering face images severely compressed or collected from unknown sources. In this paper, we propose a Face Depth Map Transformer (FDMT) to estimate the face depth map from both the original and manipulated face images. This is then treated as auxiliary information to be fused with the backbone feature by a Multi-head Depth Attention (MDA) mechanism newly designed for performance boosting.

\section{The Proposed Method}

\subsection{Overview}

Fig. \ref{fig:framework} gives an overview of our proposed method for estimating and integrating the face depth map for face manipulation detection. Given a face image for input, we partition it into a set of non-overlapping patches for patch-wise face depth estimation, where we propose a Face Depth Map Transformer (FDMT) to construct the face depth patch by patch. Next, we propose a Multi-head Depth Attention (MDA) to integrate the depth features (before the fully connected layer) extracted from the FDMT with the backbone features for enhancement. The enhanced backbone features are then fed to the rest of the backbone for classification. 

\subsection{Face Depth Map Transformer (FDMT)}

\begin{figure}%[htbp]
    \centering
    \includegraphics[width=0.45\textwidth]{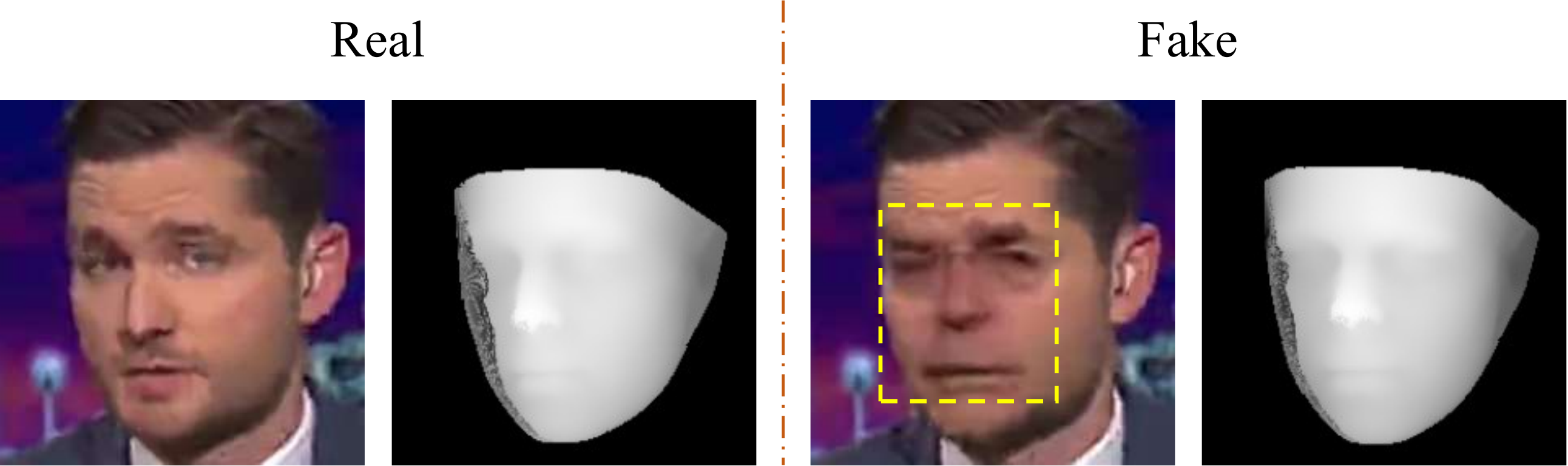}
    \caption{Examples of the estimated face depth map using PRNet \cite{feng2018joint}.  Images in the “Real” column are real face image and depth map. Images in the “Fake” column are manipulated face image and depth map, where the fake face region is bounded in yellow. The face images are selected from FaceForensics++ \cite{rossler2019faceforensics++}. \label{fig:generatedepth}}
\end{figure}

The face manipulation usually alters the original face image locally to change the face appearance. Such an operation will cause abrupt changes in the face depth map, which is unfortunately difficult to be captured by using the existing face depth map estimation schemes. Because they assume the face image is captured from a human face with relatively smooth surface, as shown in Fig. \ref{fig:generatedepth}. To make the face depth map sensitive to face manipulation, we propose here a Face Depth Map Transformer (FDMT) to estimate the face depth features patch by patch. Our FDMT is supervised by a set of ground truth face depth maps which are generated specifically for face manipulation detection. Next, we elaborate in detail on how we generate the ground truth as well as the FDMT. 

\begin{figure}%[htbp]
    \centering
    \includegraphics[width=0.45\textwidth]{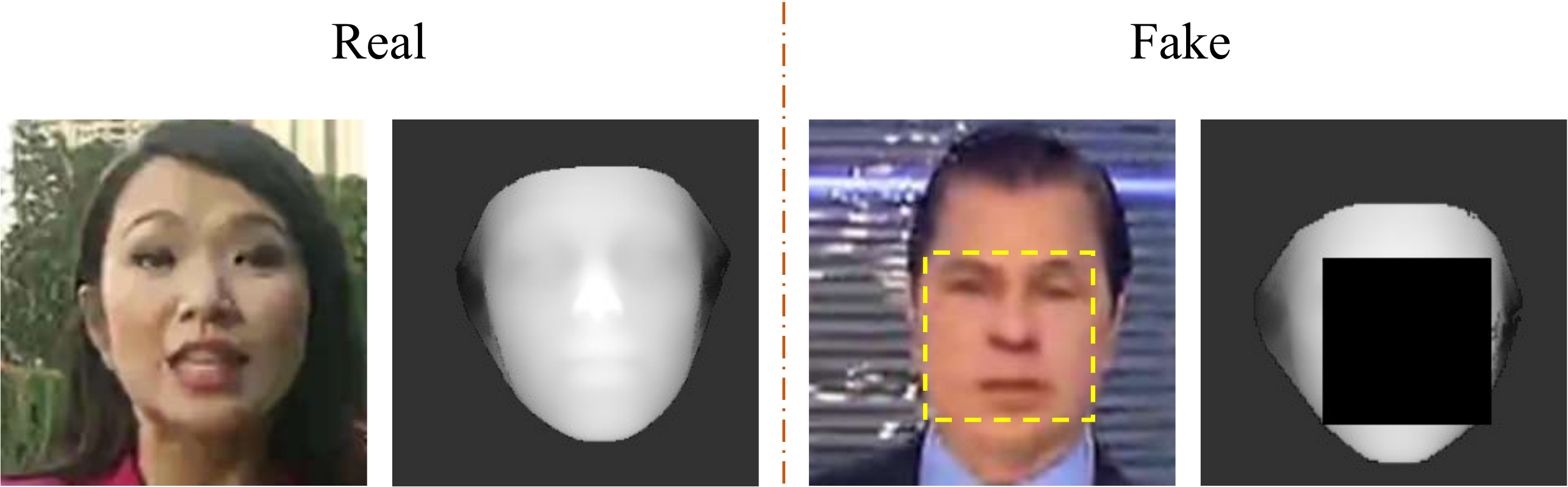}
    \caption{Examples of the ground truth face depth map. Images in the ``Real'' column are real face and the ground truth. Images in the ``Fake'' column are manipulated face image and the ground truth, where the fake face region is bounded in yellow. The face images are selected from FaceForensics++ \cite{rossler2019faceforensics++}. \label{fig:depthmap}}
\end{figure}

The ground truth face depth map should properly reflect the depth of the real face, fake face and background regions for a manipulated face image. We aware that some public face manipulation datasets may contain the masks of the fake face region for the fake face images and there are pretrained models available for depth estimation from real face images. These offer us the possibility to generate appropriate face depth maps to be served as ground truth for face manipulation detection. 

Given a face image, we first generate its face depth map based on a pretrained face depth map estimation model (PRNet) \cite{feng2018joint}. The PRNet automatically segments the image into the face region and background region, where the depth of the background region is set as 0 and the depth within the face region is represented as non-zero positive integers, and larger value means closer to the camera. Let's denote the output of PRNet as $D(x,y)$ for the pixel located at $(x,y)$ in the face image. The ground truth face depth at pixel $(x,y)$ is computed as 
\begin{equation}
     G(x,y) = \begin{cases}
                0,   &  if (x,y) \in \emph{fake face region} \\
                \Omega(D(x,y) + \lambda) ,& \emph{otherwise},
            \end{cases}
\end{equation}
where $\Omega$ is the operation to prevent overflow (i.e., set the values larger than 255 as 255) and $\lambda$ is a positive integer. As such, we have well separated depth values for different regions, where the depth of the fake face and background regions are set as 0 and $\lambda$, the depth value of the real face region is within the range from $\lambda$ to $255$. The ground truth face depth of each image patch is then computed as the average depth value within this patch based on $G(x,y)$. Fig. \ref{fig:depthmap} gives examples of our ground truth face depth map for face manipulation detection.

The structure of our FDMT is similar to ViT \cite{dosovitskiy2020vit}. We divide an input face image into a set of $P$ non-overlapping patches with positions embedded. Then, the position embedded patches are processed into $T$ transformer blocks, where each block contains two normalization layers, a multi-head attention unit and a multi layer perceptron (MLP). The output of the last transformer block is fed into a fully connected layer to produce a $P$ dimensional vector representing the depth value of each patch.  

\subsection{Multi-head Depth Attention (MDA)}
With the face depth map available, the next question is how to effectively integrate it into the backbone. A straight forward way is to concatenate it with the backbone features extracted from the RGB face image for enhancement (termed as the RGB feature for simplicity). We could also directly use it as attention to weight the RGB feature. However, both strategies ignore the correspondence between the RGB feature and the face depth map, whose correlations are not fully exploited during the integration. Here, we propose to jointly learn a depth attention by taking both the RGB feature and face depth map into consideration. 

Given the RGB feature $F_{rgb}$ extracted from the backbone and the face depth map $F_{d}$ extracted from FDMT, where $F_{rgb}$ and $F_{d}$ are with the same height and width. The RGB feature contains the color and texture information, as well as the manipulation clues in the spatial feature spaces. While the face depth map offers the corresponding depth anomalies caused by manipulation as auxiliary information. To take advantage of such correspondences, we measure the similarity between the RGB feature and the face depth map using the dot product below
\begin{equation}
     F_{rd} = (F_{d}W^{D}) \cdot (F_{rgb}W^{R}),
     \label{eq:DP_RGB}
\end{equation}
where $W^{R}$ and $W^{D}$ are the trainable weight matrices for the RGB feature and the face depth map, ``$\cdot$" is the dot product operation, and the biases are omitted for simplicity. Then, we use softmax to convert the similarity $F_{rd}$ into a depth attention by
\begin{equation}
     A_{rd} = softmax(\frac{F_{rd}}{\sqrt{d}}),
\end{equation}
where $d$ is a scaling factor equivalent to the number of channels in the RGB feature. The depth attention is eventually used to enhance the RGB feature by
\begin{equation}
     F'_{rgb} = A_{rd} \odot (F_{rgb}W^{V}),
\end{equation}
where $W^{V}$ is a trainable weight matrix, ``$\odot$" is the element-wise multiply operation. The backbone feature is then enhanced below by fusing the RGB features before and after the depth attention 
\begin{equation}
    F_{en} = F_{rgb} + \text{MLP}(F'_{rgb}+F_{rgb}).
\end{equation}

Next, we adopt the multi-head strategy \cite{Vaswani2017attention} on our depth attention to achieve an $l$-head depth attention, which is given as 
\begin{multline}
\label{eq:multihead}
    \text{MultiHead}(F_{d}W^{D}, F_{rgb}W^{R}, F_{rgb}W^{V}) = \\
        \text{Concat}(d_{1},d_{2},\dots,d_{l})W^{O},
\end{multline} 
where $d_i$ refers to the $i$-th head with input being $F_{d}W^{D}$, $F_{rgb}W^{R}$, $F_{rgb}W^{V}$ and $F_{en}$ being the output, ``Concat" is the concatenation operation, and $W^{O}$ is a weight matrix to aggregate the outputs of different heads. Note that the trainable matrices for $F_{d}$ and $F_{rgb}$ are not shared among different heads.  

\subsection{Loss Function}
We adopt the SSIM loss and the MSE loss to evaluate the similarity between the estimated and the ground truth face depth map. The SSIM loss is formulated as 
\begin{equation}
    \mathcal{L}_{ssim} = \frac{(2\mu_{a}\mu_{b}+c_{1})(2\sigma_{ab}+c_{2})}{(\mu_{a}^{2}+\mu_{b}^{2}+c_{1})(\sigma_{a}^{2}+\sigma_{b}^{2}+c_{2})},
\end{equation}
where $\mu_{a}$, $\mu_b$ represent the mean of the estimated and ground truth face depth map; $\sigma_{a}^{2}$, $\sigma_{b}^{2}$ denote the corresponding variance; $\sigma_{ab}$ is the covariance; $c_{1}$ and $c_{2}$ are small positive integers to avoid the division by zero. The MSE loss is given by \begin{equation}
    \mathcal{L}_{patch\_mse} = \sum_{i=1}^{M}\sum_{p=1}^{P}{||a_{i,p}-b_{i,p}||_{2}},
\end{equation}
where $a_{i,p}$ and $b_{i, p}$ are the depth values of the $p$-th patch in the estimated and ground truth face depth map for the $i$-th training sample, respectively.

The total loss is computed as

\begin{equation}
    \label{eq:loss}
    \mathcal{L}_{total}=\mathcal{L}_{c} + \alpha\cdot\mathcal{L}_{ssim} + \beta\cdot\mathcal{L}_{patch\_mse},
\end{equation}
where $\mathcal{L}_{c}$ is the backbone loss for image classification, $\alpha$ and $\beta$ are the weights to balance different loss terms.

\section{Experiments}

\subsection{Setup}

\subsubsection{Dataset}~We use two large-scale face manipulation datasets: FaceForensics++ (FF++) \cite{rossler2019faceforensics++} and Celeb-DF \cite{li2020celebdf} for experiments. The FF++ dataset contains 1,000 real videos, with 720 videos for training, 140 videos for validation and 140 videos for testing. Each video has four versions using different manipulation methods, which are DeepFakes (DF) \cite{Tora2018deepfakes}, Face2Face (F2F) \cite{thies2016face2face}, FaceSwap (FS) \cite{Kowalski2018faceswap} and Neural Textures (NT) \cite{thies2019deferred}. Besides, each video has three compression levels, which are RAW, High Quality (c23) and Low quality (c40). The Celeb-DF dataset includes 590 raw videos collected from YouTube and 5639 corresponding face manipulated videos, which covers refined fake face videos from different genders, ages and races with similar quality to those transmitted in real world scenarios. 

\subsubsection{Evaluation Metrics}~We use the detection accuracy (ACC) and Area under the Curve (AUC) for evaluation, which are two common indicators in literature for evaluating face manipulation detection schemes. 

\subsubsection{Implementation Details}~We take Xception \cite{chollet2017xception} as a backbone by default to evaluate the performance of our proposed method, where we integrate our Multi-head Depth Attention between the seventh to the eighth block of the Xception. For each image, we follow the suggestion given in \cite{rossler2019faceforensics++} to automatically and conservatively crop the facial area into a square, which is then resized to $224 \times 224$ for training and testing. We use the masks of the fake face region provided by FF++ to generate the ground truth face depth map with $\lambda=50$, which is then normalized to $[0,1]$ for training. Our FDMT contains $T=12$ blocks with eight attention heads for each block, where the input image is partitioned into $P=14\times14$ patches. Both the values of $\alpha$ and $\beta$ are set as $0.7$ for the total loss. The model is trained with Adam optimizer \cite{Kingma2015AdamAM} with a learning rate $3\times10^{-4}$ and a weight decay $10^{-4}$. We train our model on a RTX 3090 GPU with a batch size of $32$.

\subsection{Comparisons}

In this section, we compare our method with the existing image level face manipulation detection methods. We train our model on FF++ and test it on FF++ for intra-database evaluation and Celeb-DF for cross-database evaluation. The performance of the existing works are all duplicated from literature.%We also explore the effects of different backbones and different insertion position on deepfake detection performance.

% (Section \ref{sec:intra})  (Section \ref{sec:cross})

% FF++数据集c40压缩质量各操作手段子集上训练测试结果
\begin{table} %[htbp]
    \centering
    \resizebox{1\columnwidth}{!}{
        \begin{tabular}{cccccc}
            \hline
            Method               & DF      & F2F      & FS       & NT   \\ 
            \hline
             Xception  \cite{chollet2017xception}            &  92.04   &  87.03  &  90.31    &  70.43 \\
             FaceForensics++ \cite{rossler2019faceforensics++}      &  92.43  &  80.21   &  88.09   &  56.75  \\
             Mesonet   \cite{afchar2018mesonet}            &  87.27  &  56.20   &  61.17   &  40.67  \\
             F$^{3}$Net     \cite{qian2020thinking}         &  93.06  &  81.48   &  89.58   &  61.95  \\
             ADD-ResNet50   \cite{woo2022add}       &  \textbf{95.50}  &  85.42   &  \textbf{92.49}   &  68.53  \\
             Ours (Xception)       &  94.01  &  \textbf{88.78}   &  91.97   &  \textbf{72.32}  \\ 
            \hline
        \end{tabular}
    }
    \caption{The ACC (\%) of different schemes on the four FF++ subsets under low quality setting (compression quality: c40).}
    \label{tab:FF++}
\end{table}

\subsubsection{Intra-database Evaluation}
We compare our method with the existing schemes in terms of ACC on low quality compression setting (c40) in FF++. In particular, we train and test our model on the four subsets of the low quality compressed FF++ for different manipulation methods. The detection results of different approaches are shown in Table \ref{tab:FF++}. It can be seen that, compared with the original Xception, our proposed method achieves $1.97\%$, $1.75\%$, $1.66\%$, and $1.89\%$ performance gain on DF, F2F, FS, and NT, respectively. Compared with the ADD-Resnet50 \cite{woo2022add} which is state-of-the-art method specifically proposed for low quality compressed fake face image detection, our method achieves comparable performance on the DF and FS subsets, and achieves $3.36\%$ and $3.79\%$ higher accuracy on the F2F and NT subsets, respectively. We would like to mention that the results of the ADD-Resnet50 are based on knowledge distillation. It requires a teacher model to be pre-trained on high-quality face images, which is not purely trained on low-quality compressed face images as what we have done for our proposed method. 

\begin{table}%[htbp]
    \centering
    \resizebox{0.95\columnwidth}{!}{
        \begin{tabular}{ccc}
        \hline
        Method         & FF++           & Celeb-DF       \\ 
        \hline
        Two-stream  \cite{Zhou2017two}   & 70.10          & 53.80          \\
        Two-branch  \cite{masi2020two}    & 93.20          & 73.40          \\
        Meso4   \cite{afchar2018mesonet}       & 84.70          & 54.80          \\
        MesoInception4  \cite{afchar2018mesonet} & 83.00          & 53.60          \\
        FWA    \cite{li2018exposing}        & 80.10          & 56.90          \\
        DSP-FWA   \cite{li2018exposing}     & 93.00          & 64.60          \\
        %Capsule   \cite{nguyen2019capsule}     & 96.60          & 57.50          \\
        Xception-c23 \cite{rossler2019faceforensics++}  & 99.70              & 65.30          \\
        Multi-task \cite{Nguyen2019Multi}    & 76.30          & 54.30          \\
        F$^{3}$-Net \cite{qian2020thinking}         & 97.97 & 65.17          \\
        Mutil-attention \cite{zhao2021multi}          & \textbf{99.80}        & 67.44 \\
        PD-Xception   \cite{zhang2022patch}          & 98.71         & 74.22         \\
        Ours (Xception)           & 97.29          & \textbf{80.88} \\
        \hline
        \end{tabular}
    }
    \caption{The AUC(\%) of different schemes for cross-database evaluation.}
    \label{tab:cross-database}
\end{table}

\subsubsection{Cross-database Evaluation\label{sec:cross}}
By following the suggestion given in most of the existing works, we train our model on the whole training set of FF++ (c23) dataset and test it on the Celeb-DF test set for cross-database evaluation. Table \ref{tab:cross-database} shows the AUC of different schemes, where the performance on the test set of FF++ (c23) is also given for reference. It can be seen from Table \ref{tab:cross-database} that our method is significantly better than the existing schemes for cross-database evaluation with over 6.66\% higher AUC when tested on Celeb-DF. On the other hand, our method achieves satisfactory performance on the test set of FF++ (c23) with AUC of 97.29\%. 

\subsection{Sensitivity of Multi-head Depth Attention (MDA)}
As shown in Fig. \ref{fig:framework}, our proposed MDA has to be integrated into the backbone for feature enhancement. In this section, we test the sensitivity of our MDA on the F2F subset of FF++ (c40) by integrating it with the feature maps of different blocks in the default Xception backbone. We denote the 13 blocks in Xception as B1, B2, ... , B13 from input to output. We integrate the MDA with the feature maps extracted from three different blocks: B3, B7, and B11, the results of which are given in Table \ref{tab:insertion}. It can be seen that all the integration strategies improve the detection accuracy and the improvement does not vary much. While the integration with the features extracted from the center block (i.e., B7) achieves the best. 

% face2face ACC
\begin{table}%[htbp]
    \centering

    \resizebox{0.70\columnwidth}{!}{
    \begin{tabular}{ccc}
    \hline
    Different blocks for integration & ACC        \\ 
    \hline
     -           & 87.03        \\
    B3    & 88.45       \\
    B7    & \textbf{88.78}          \\
    B11  & 88.25   \\ 
    \hline
    \end{tabular}
    }
    \caption{The performance of integrating the MDA into different blocks in Xception.}
    \label{tab:insertion}
\end{table}

\subsection{Generalization Ability against Different Backbones}
In this section, we integrate our proposed method with three popular face manipulation detection backbones including Xception \cite{chollet2017xception}, Resnet50 \cite{he2016resnet}, and EfficientNet \cite{tan2019efficientnet}, where the EfficientNet is the runner up in the facebook Deepfake Detection Challenge \cite{dolhansky2019deepfake}. We train all the models on the whole training set of FF++ (c40), and test them separately on the test set of FF++ (c40) and Celeb-DF. We conduct the MDA integration with the feature maps extracted right after the seventh block, the third layer, and the twelfth block in the Xception, Resnet50, and EfficientNet. 

Table \ref{tab:backbone} gives the performance before and after using our proposed method for integration. It can be seen that, regardless of the backbones, our proposed method is able to boost the performance, especially for the cross-database scenarios, where the performance gain for Xception, ResNet50, and EfficientNet is $5.34\%$, $5.75\%$, and $2.39\%$ in terms of AUC, respectively. These indicate the good generalization ability of our proposed method for performance boosting on different backbones.

\begin{table}%[htpb]
\centering
\resizebox{0.95\columnwidth}{!}{
    \begin{tabular}{cccc}
        \hline
        \multirow{2}{*}{Backbone} &FF++ (c40) & Celeb-DF \\ \cline{2-3} 
                                  & ACC                & AUC      \\   
        \hline 
        Xception\cite{chollet2017xception}                                         & 81.33               & 63.13          \\
        Ours (Xception)                                                            & \textbf{82.52}  & \textbf{68.47} \\
        ResNet50\cite{he2016resnet}                                                & 79.65                   & 49.50          \\
        Ours (ResNet50)                                                            & \textbf{80.34}   & \textbf{55.25} \\
        EfficientNet\cite{tan2019efficientnet}                                     & 81.77                     & 62.89          \\
        Ours (EfficientNet)                                                        & \textbf{81.81}   & \textbf{65.28} \\ 
        \hline
    \end{tabular}
    }
\caption{The performance of different backbones before and after integrating our proposed method.}
\label{tab:backbone}
\end{table}

\subsection{Ablation Study}

To verify the effectiveness of our proposed Face Depth Map Transformer (FDMT) and Multi-head Depth Attention (MDA), we evaluate them separately in this section, where all the models are trained and tested on the F2F subset in FF++ with the backbone fixed as Xception, and we take the backbone feature extracted from the seventh block in Xception for fusion or attention when necessary.  

\subsubsection{Effectiveness of FDMT}
To demonstrate the effectiveness of FDMT, we conduct two additional experiments here: 1) we concatenate the face depth maps extracted from FDMT with the backbone feature, and 2) we concatenate the face depth maps extracted from an existing face depth estimation model PRNet \cite{feng2018joint} with the backbone feature. The concatenated features are then passed through a 1x1 convolutional layer to obtain fused features for the subsequent Xception blocks. Table \ref{tab:Ab_depth} gives the ACC of the aforementioned experiments as well as those of the Xception backbone. It can be seen that both the two experiments achieve higher ACC compared with using the Xception backbone only. This indicates that the face depth map is indeed helpful for the face manipulation detection task. By simply concatenating an existing face depth map with the backbone feature, we are able to achieve 0.65\% improvement in ACC. While the face depth map estimated using our proposed FDMT is more effective for performance boosting, with 1.2\% of improvement in ACC compared with using the Xception backbone.

% Effectiveness of FDMT
\begin{table}%[htbp]
    \centering
    \resizebox{0.75\columnwidth}{!}{
        \begin{tabular}{cccc}
        \hline
        Depth Map & Fusion & ACC  \\ 
        \hline
        -         & -      & 87.03  \\
        PRNet\cite{feng2018joint}     & concat & 87.68 \\
        FDMT      & concat & \textbf{88.23}\\
        \hline
        \end{tabular}
    }
    \caption{Ablation study for the FDMT.}
    \label{tab:Ab_depth}
\end{table}

\begin{figure}%[htbp]
    \centering
    \includegraphics[width=0.45\textwidth]{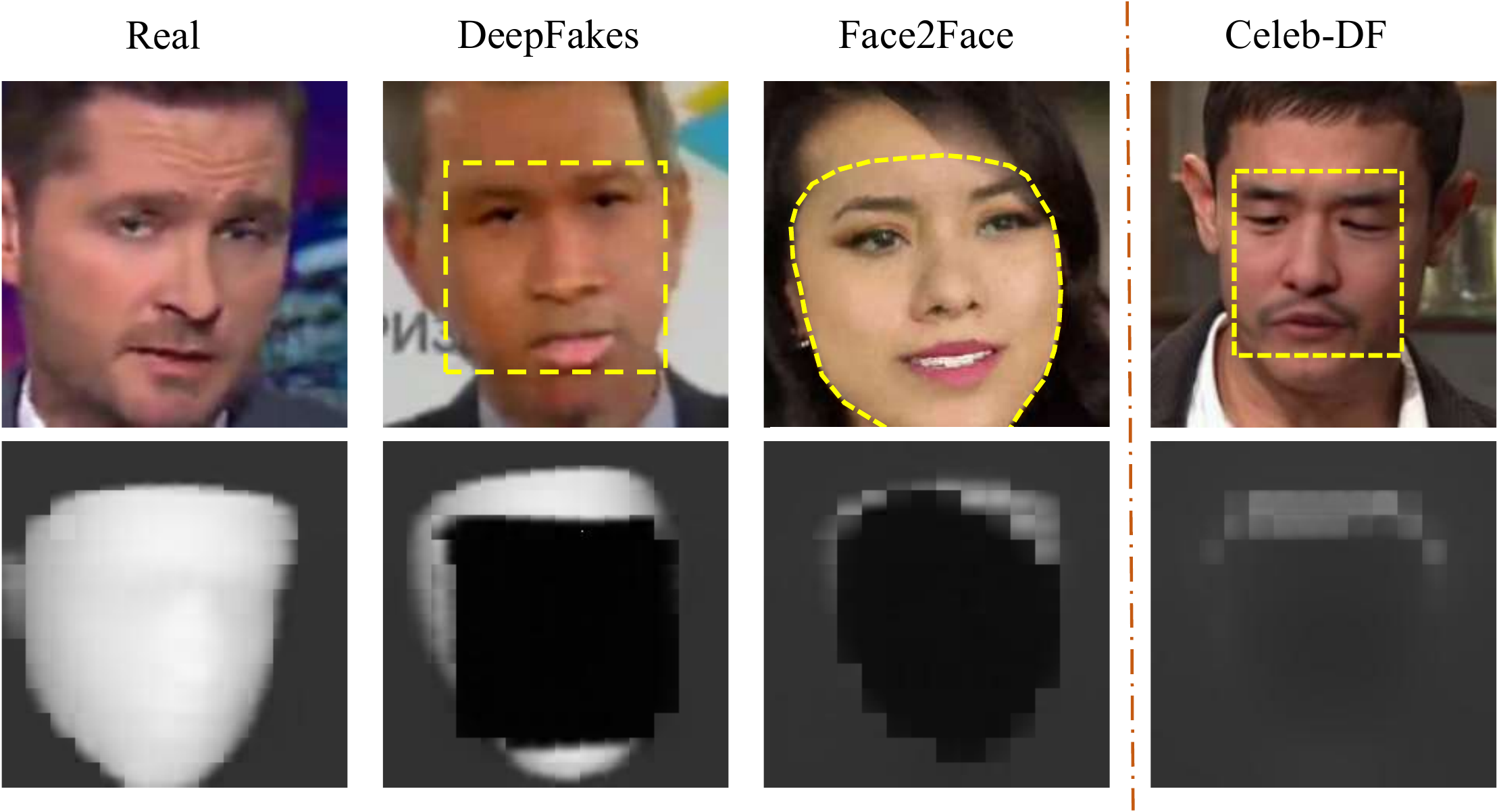}
    \caption{Examples of the face depth maps estimated using our proposed FDMT. The first row shows the face images selected from different databases with the fake face regions bounded in yellow, while second row gives the corresponding face depth maps.}
    \label{fig:depth_generate}
    % Images in the ”Real” column are realistic face and depth map. Images in other columns are manipulated faces and corresponding face depth maps.
\end{figure}

Fig. \ref{fig:depth_generate} gives some examples of the estimated face depth map from real and fake face images using the the proposed FDMT. The images on the first three columns are from the FF++ dataset, and the image on the fourth column is selected from Celeb-DF. It can be seen that our FDMT can effectively estimate the face depth, which is able to capture the anomaly in face depth caused by face manipulation for face images from different sources. This further demonstrates the ability of our FDMT to extract robust face depth features for face manipulation detection.

%Face manipulation usually alters the original face image, which includes changing depth information. Compared with the existing depth estimation algorithms, such as PRNet\cite{feng2018joint} , it only considers that the input image is a face with a relatively smooth surface, and at the beginning of the design, it does not have the ability to distinguish between the real face area and the fake face area. The FDMT proposed for the face manipulation detection task is sensitive to face manipulation and can detect the depth changes caused by face manipulation.

\subsubsection{Effectiveness of MDA}

To verify the effectiveness of our proposed MDA, we conduct two more experiments here: 1) we replace our FDMT with the PRNet for face depth estimation, 2) we incorporate the popular multi-head self-attention (MSA) \cite{Vaswani2017attention} with the backbone feature without using the face depth map. Table \ref{tab:DepthAttention} gives the ACC of face manipulation detection for different models. It can be seen that, by simply using the MSA, we are able to achieve around 1\% higher in ACC compared with using the Xception backbone only. Our MDA is able to further increase the ACC which are 0.17\% and 0.7\% higher than that of using the MSA by depth attention based on PRNet and our propose FDMT, respectively. Compared with the results of directly concatenating the face depth map (see Table \ref{fig:depth_generate}), our MDA achieves more performance gain with around 0.5\% improvement in ACC for both the face depth maps estimated using the PRNet and our proposed FDMT. These results indicate that the multi-head based attention mechanism works for the task of face manipulation detection, and our MDA is superior to the existing MSA  with the help of face depth map. 

% face2face ACC AUC
\begin{table}%[htbp]
    \centering
    \resizebox{0.75\columnwidth}{!}{
        \begin{tabular}{cccc}
        \hline
        Depth Map & Attention & ACC \\ \hline
        -         & -         & 87.03 \\
        -     & MSA           & 88.08 \\
        PRNet\cite{feng2018joint}     & MDA       & 88.25\\
        FDMT      & MDA       & \textbf{88.78}\\ \hline
        \end{tabular}
    }
    \caption{Ablation study for the MDA.}
    \label{tab:DepthAttention}
\end{table}

\begin{figure}[t]%[htbp]
    \centering
    \includegraphics[width=0.47\textwidth]{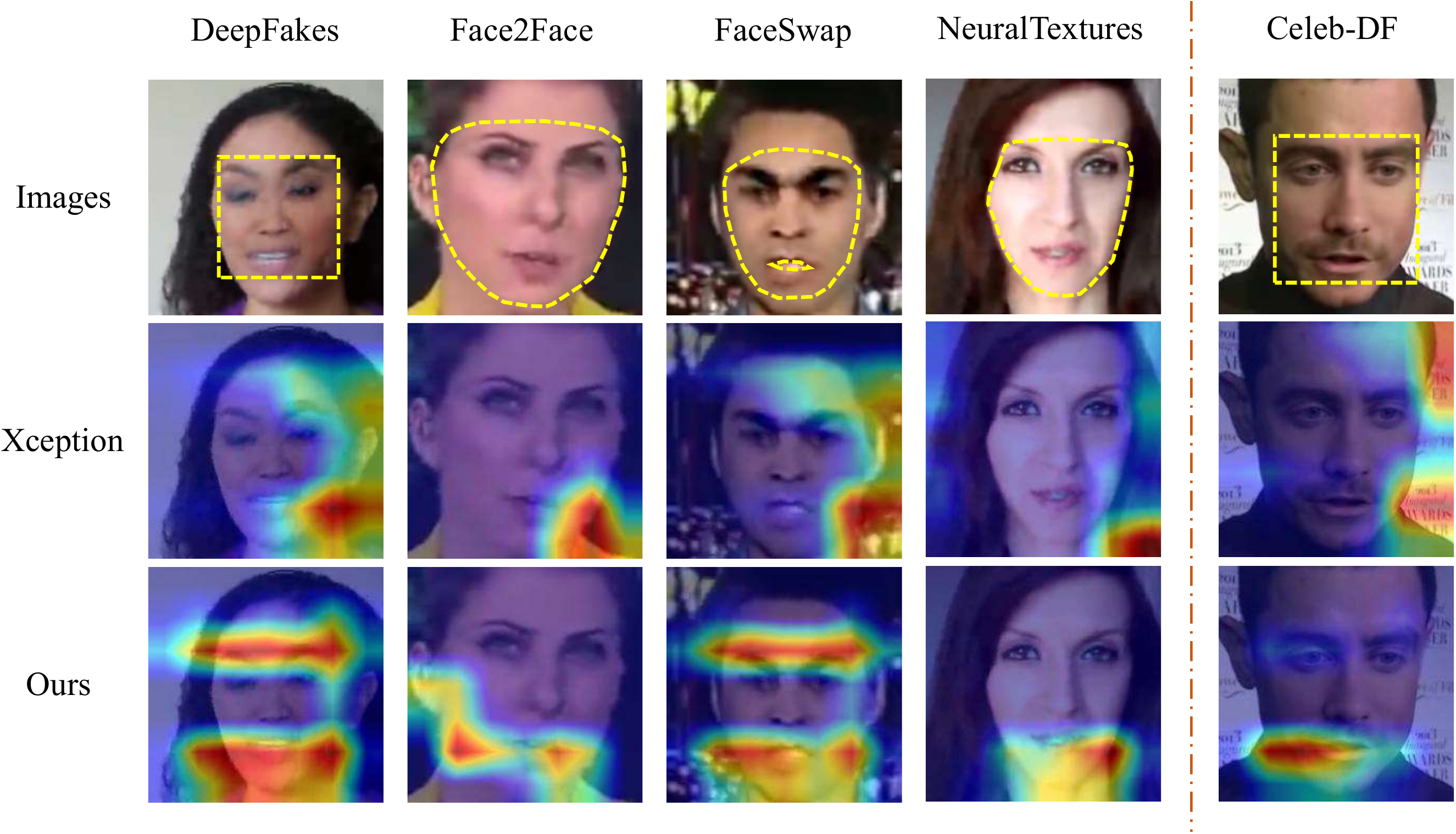}
    \caption{Visulization of the features before and after using our proposed method on the Xception backbone. From top to bottom: manipulated face images with the fake face region bounded in yellow (first row), visualization of the features before (second row) and after (third row) using our proposed method. The first four columns give examples from different subsets in the FF++ dataset, and the fifth column illustrates an example from the Celeb-DF dataset.}
    \label{fig:GradCam}
\end{figure}

Next, we visualize the Gradient-weighted Class Activation Mapping (Grad-CAM) \cite{selvaraju2017grad} of the backbone features before and after the integration of our proposed method, as shown in Fig. \ref{fig:GradCam}. It can be seen that, by using our proposed method, the backbone feature is able to focus more on the fake face region, which is helpful for accurate face manipulation detection.

\section{Conclusion}

In this paper, we explore the possibility of using the face depth map to facilitate the face manipulation detection. To extract representative face depth maps from the manipulated face images, we design a Face Depth Map Transformer (FDMT) to estimate the face depth features patch by patch, which is effective in capturing the local depth anomaly created due to the manipulation. To appropriately integrate the face depth map, we further propose a Multi-head Depth Attention (MDA) mechanism to enhance the backbone features via a depth attention which is computed by the scale dot product between the face depth map and the backbone feature of the RGB face image. Experimental results indicate that our proposed method is particularly helpful in cross-database scenarios with over 6.6\% higher AUC than the existing schemes, which also works well for detecting low quality compressed manipulated face images.       
\bibliography{reference}

% \appendix

% \section{Acknowledgments}

% \bigskip

\end{document}